# Refining reasoning in qualitative probabilistic networks

Simon Parsons*
Advanced Computation Laboratory
Imperial Cancer Research Fund
P.O. Box 123
Lincoln's Inn Fields
London WC2A 3PX, UK

## Abstract

In recent years there has been a spate of papers describing systems for probabilisitic reasoning which do not use numerical probabilities. In some cases the simple set of values used by these systems make it impossible to predict how a probability will change or which hypothesis is most likely given certain evidence. This paper concentrates on such situations, and suggests a number of ways in which they may be resolved by refining the representation.

## 1 INTRODUCTION

In the past few years there has been considerable interest in qualitative approaches to reasoning under uncertainty—approaches which do not make use of precise numerical values of the type used by conventional probability theory. These approaches range from systems of argumentation (Benferhat, Dubois, & Prade 1993; Darwiche 1993; Fox, Krause, & Ambler 1992) to systems for nonmonotonic reasoning (Goldszmidt 1992) and abstractions of precise quantitative systems (Druzdzel & Henrion 1993; Wellman 1990). Qualitative abstractions of probabilistic networks, in particular, have proved popular, finding use in areas in which the full numerical formalism is neither necessary nor appropriate. Applications have been reported in explanation (Henrion & Druzdzel 1990), diagnosis (Darwiche & Goldszmidt 1994; Henrion et al. 1994), engineering design (Michelena 1991), and planning (Wellman 1990).

In qualitative probabilistic networks (QPNs), the focus is rather different from that of ordinary probabilistic systems. Whereas in probabilistic networks (Pearl

*Current address: Department of Electronic Engineering, Queen Mary and Westfield College, Mile End Road, London E1 4NS, UK

1988) the main goal is to establish the probabilities of hypotheses when particular observations are made, in qualitative systems the main aim is to establish how values change rather than what the values are. Since the approach is qualitative, the size of the changes are not required. The only consideration is whether a given change is positive, written as [+], negative [−], or zero [0]. In some cases it is not possible to resolve the change with any precision so that its value remains unknown, and it is written as [?]. Clearly this information is rather weak, but as the applications show it is sufficient for some tasks. Furthermore, reasoning with qualitative probabilities is much more efficient than reasoning with precise probabilities, since computation is quadratic in the size of the network (Druzdzel & Henrion 1993), rather than NP-hard (Cooper 1990).

The popularity of qualitative probabilistic networks prompted work on abstractions of other uncertainty handling formalisms (Parsons 1995b; 1995a; Parsons & Mamdani 1993), providing what is essentially a generalisation of the approach provided by qualitative probabilistic networks (Wellman 1993) to what are termed qualitative certainty networks (QCNs). The approach uses techniques from qualitative reasoning (Bobrow 1984) to determine the behaviour of the formalisms. Using this approach it is possible to propagate qualitative probability, possibility (Dubois & Prade 1988; Zadeh 1978) and Dempster-Shafer belief (Shafer 1976) in a uniform way.

There are two major problems with both qualitative probabilistic and certainty networks, which are related to their level of abstraction. The first is that they cannot always predict which of a pair of hypotheses is most likely given certain evidence. The second is that if a particular hypothesis is influenced by two pieces of evidence, one of which makes it more likely and one of which makes it less likely, then if both are observed, it is not possible to tell what the change in probability of the hypothesis is. This paper gives a number of ways in which this problem may be tackled.

## 2  BASIC NOTIONS

Both QPNs and QCNs are built around the notion of influences between variables represented by nodes in a graph. In this section we introduce the basic notions behind both, and show how, in the binary case, they are equivalent so that the results given later in the paper hold equally for both approaches. The description of a QPN is that given by Druzdzel and Henrion (1993) and is marginally adapted to fit in with the notation of QCNs. Formally, a QPN is a pair $G = (V, Q)$, where $V$ is a set of variables or nodes in the graph, denoted by capital letters, and $Q$ is a set of qualitative relations among the variables. There are two types of qualitative relations in $Q$, *influences* and *synergies*, but here we are only interested in influences. These define the sign of the direct influence between variables and correspond to arcs in a probabilistic network.

**Definition 1 (qualitative influence)** *We say that $A$ positively influences $C$, written $S^{[+]}(A, C)$, iff for all values $a_1 > a_2$, $c_0$, and $X$, which is the set of all of $C$'s predecessors other than $A$:*

$$p(c \geq c_0 \,|\, a_1, X) \geq p(c \geq c_0 \,|\, a_2, X)$$

where $a_i$ and $c_j$ are the possible values of $A$ and $C$. This definition expresses the fact that increasing the value of $A$ makes higher values of $C$ more probable. *Negative qualitative influence*, $S^{[-]}$, and *zero qualitative influence*, $S^{[0]}$, are defined analogously by substituting $\leq$ and $=$ respectively for $\geq$. To allow belief propagation it is necessary to propagate qualitative changes in value in both directions. This is made possible by the following theorem (Milgrom 1981):

**Theorem 2 (symmetry of influences)** $S^{[\delta]}(A, C)$ *implies* $S^{[\delta]}(C, A)$.

The impact of evidence on a given node can be calculated by taking the sign of the change in probability at the evidence node and multiplying it by the sign of every link in the sequence of links that connect it to the node of interest. To see how this works, consider the example in Figure 1 which is an adaptation of fragment of the car diagnosis network of Henrion *et al.* (1994). If we observe that the radio is dead, so that the probability of the radio being ok decreases, $p(radio\ ok) = [-]$, and we want to know the impact of this on the state of the battery we calculate the effect as $[-] \otimes [+] \otimes [+]$. With the definition of sign multiplication in Table 1 this gives a change in $p(battery\ good)$ of $[-]$. If we also observed that the lights were not ok, and wanted to assess the impact of both pieces of evidence on the probability that the battery was good, we would establish the two individual effects and sum

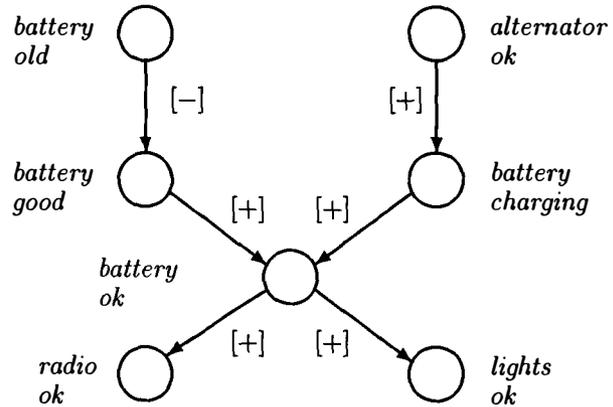

Figure 1: Part of a car diagnosis network

them using $\oplus$ (Table 2). QCNs are very similar, the main differences being that they are abstractions of possibilistic and Dempster-Shafer belief networks as

Table 1: Sign multiplication.

| $\otimes$ | [+] | [0] | [-] | [?] |
|---|---|---|---|---|
| [+] | [+] | [0] | [-] | [?] |
| [0] | [0] | [0] | [0] | [0] |
| [-] | [-] | [0] | [+] | [?] |
| [?] | [?] | [0] | [?] | [?] |

Table 2: Sign addition.

| $\oplus$ | [+] | [0] | [-] | [?] |
|---|---|---|---|---|
| [+] | [+] | [+] | [?] | [?] |
| [0] | [+] | [0] | [-] | [?] |
| [-] | [?] | [-] | [-] | [?] |
| [?] | [?] | [?] | [?] | [?] |

well as probabilistic networks, and that, in general, the qualitative influences between variables need more than one sign to define them. Formally, a QCN is a pair $G = (V, Q)$, where $V$ is a set of variables or nodes in the graph, once again represented by a capital letter, and $Q$ is a set of sets of qualitative relations among the values of the variables. The qualitative relations are based upon the derivatives that relate the different values of the variables together. In the case of a probabilistic QCN we have:

**Definition 3 (qualitative derivative)** *The qualitative derivative $\left[\frac{dp(c_1)}{dp(a_1)}\right]$ relating the probability of $C$ taking value $c_1$ to the probability of $A$ taking value $a_1$ has the value $[+]$, if, for all $a_2$ and $X$:*

$$p(c_1 \,|\, a_1, X) \geq p(c_1 \,|\, a_2, X)$$

Derivatives with values $[-]$ and $[0]$ are defined by replacing $\geq$ with $\leq$ and $=$. The similarity of Definitions 1 and 3 means that it is no surprise to find that the propagation of qualitative changes in the value of variables is once again performed using $\otimes$ and $\oplus$. If we write the qualitative value of the change in probability of variable $A$ taking value $a_1$ as $[\Delta p(a_1)]$ then:

$$[\Delta p(c_1)] = \left[\frac{dp(c_1)}{dp(a_1)}\right] \otimes [\Delta p(a_1)]$$

and the overall effect of multiple changes is calculated using $\oplus$. Now, a link in a QCN is normally specified by a number of qualitative values—one for each relevant derivative, which means one for every pair of values, one from each variable, of the two variables joined by the link—while, as stated above a link in a QPN is completely specified by a single value. This means that despite their similarities the two methods differ in their representation of the same information. However, when we have a binary probabilistic link, things are rather simpler. In this case, the condition on the derivative relating $p(c)$ to $p(a)$ being $[+]$ is:

$$p(c \mid a, X) \geq p(c \mid \neg a, X)$$

which means that if $\left[\frac{dp(c)}{dp(a)}\right] = [+]$ it is necessarily the case that $\left[\frac{dp(c)}{dp(\neg a)}\right] = [-]$ and, furthermore the relation between $p(c)$ and $p(\neg c)$ makes it necessary that $\left[\frac{dp(\neg c)}{dp(a)}\right] = [-]$ and $\left[\frac{dp(\neg c)}{dp(\neg a)}\right] = [+]$. These derivatives express exactly the same as a positive qualitative influence between binary valued $A$ and $C$ (if $a > \neg a$ and $c > \neg c$)—as $a$ becomes more probable, $c$ becomes more probable as well. It is also the case that in this binary case $\left[\frac{dp(c)}{dp(a)}\right] = \left[\frac{dp(a)}{dp(c)}\right]$ so that links are symmetric. The upshot of this is that binary probabilistic QCNs are equivalent to binary QPNs, and their links can be summarised by a single qualitative value which is the qualitative derivative relating the "true" values of the links that the node connects. Thus, by investigating binary probabilistic QCNs we simultaneously develop results applicable to work involving QPNs, and this is what will be undertaken in the rest of this paper, refering to both systems simultaneously as QP/CNs.

Consider the QP/CN in Figure 2 which gives some information about illness and employment. If I become ill it is more likely than before that I will lose my job and more likely that I will have to go to hospital. In addition, if it is discovered that I am not qualified, it becomes more likely that I will lose my job. However, regular exercise makes it more likely that I will be fit, and staying fit makes it less likely that I will end up in hospital. Consider further that it is known that I am ill. Propagating the effect of this information in our QP/CN tells us that both ending up in hospital and losing my job become more likely since both hypotheses have value $[+]$. Thus it is not clear which is most likely. Distinguishing which of these competing hypotheses is more likely is the first problem that we will address in this paper. Also consider what happens if I both become ill and exercise—the first makes hospital more likely, the second makes it less likely. Propagation of the effects of both these peices of information in the QP/CN yields a value of $[?]$ for the change in probability of the hypothesis so that we cannot say how it will change. This problem of competing influences is the second point we will address.

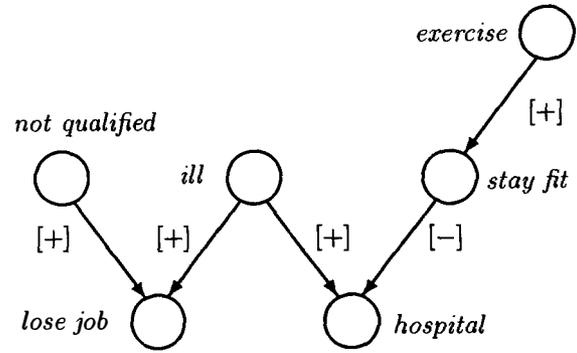

Figure 2: Some recent events (I).

## 3  DISTINGUISHING TRUTH

The problem that we face is one of over-abstraction, and it is one well known in qualitative physics. One of the methods used to handle it is the use of landmarks (Bobrow 1984), that is distinguishing important values of variables and calculating changes relative to them. Currently QP/CNs handle links that cause a change in the descendant when the parent changes. If we distinguish 1 and 0, which are the obvious landmarks for probability theory, we can also distinguish increases to 1, decreases to 0, and links in which the change in the descendent is to a value of 1 or 0. More formally we denote an increase to 1 as $[\bar{\uparrow}]$, a decrease to zero as $[\underline{\downarrow}]$, and introduce a new kind of qualitative influence based on the absolute value of the conditionals:

**Definition 4 (categorical influence)** *We say that $A$ positively categorically influences $C$, written $S^{[++]}(A,C)$, iff for all $X$, $p(c|a,X) = 1$.*

which it is easy to see will ensure that $p(c \mid a) = 1$ so that whenever $p(a)$ increases to 1, $p(c)$ increases to 1. We can similarly define an negative categorical influence $S^{[--]}(A,C)$ which ensures that whenever $p(a)$ increases to 1, $p(c)$ decreases to 0 by making $p(c \mid a, X) = 0$. Clearly neither of these types of influence is symmetric since, for instance, the fact that $p(c)$ is 1 whenever $p(a)$ is 1 does not mean that $p(a)$ is 1 whenever $p(c)$ is 1. Thus a categorical influence between $A$ and $C$ does not imply a categorical influence between $C$ and $A$. In order to propagate values with the new influences we need to extend the definition of $\otimes$ to that in Table 3.

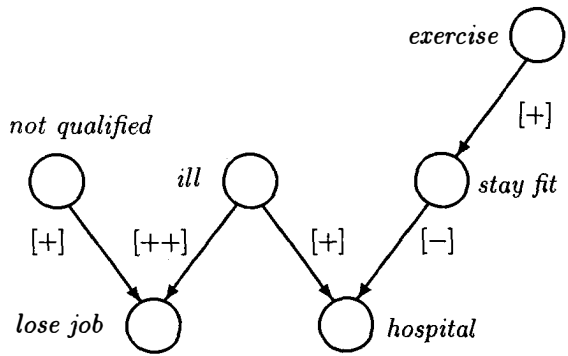

Figure 3: Some recent events (II).

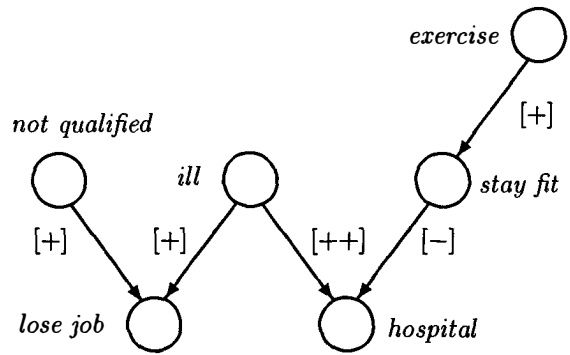

Figure 4: Some recent events (III)

Table 3: New sign multiplication.

| $\otimes$ | [++] | [+] | [0] | [−] | [−−] | [?] |
|---|---|---|---|---|---|---|
| [↑̄] | [↑̄] | [+] | [0] | [−] | [↓̲] | [?] |
| [+] | [+] | [+] | [0] | [−] | [−] | [?] |
| [0] | [0] | [0] | [0] | [0] | [0] | [0] |
| [−] | [−] | [−] | [0] | [+] | [+] | [?] |
| [↓̲] | [−] | [−] | [0] | [+] | [+] | [?] |
| [?] | [?] | [?] | [0] | [?] | [?] | [?] |

This is sufficient to solve the problem of competing hypotheses in the special case that one of the hypotheses is connected to an observed event by a chain of categorical influences. For example, consider Figure 3 in which the representation of my recent history is updated to make it more realistic. Here, when it is known that I am ill, we find that $\Delta p(hospital) = [+]$, while $\Delta p(lose\ job) = [\bar{\uparrow}]$, so that we know that it is at least as likely that I will lose my job as it is that I will have to go to hospital.

To combine categorical and qualitative influences, we need to define a new version of $\oplus$. Initially it might seem as though we have 16 possible cases to consider—every possible combination of the two types of influence. However, two are ruled out by the restrictive probability distribution that comes with a categorical influence:

**Property 5 (restricted representation)** *If there is an influence $S^{[++]}(A, C)$ then there can be no influence $S^{[--]}(X, C)$ and vice-versa.*

**Proof:** For $S^{[++]}(A, C)$ we require $p(c \mid a, X) = 1$ and thus $p(c \mid a, x) = 1$. For $S^{[--]}(X, C)$ we require $p(c \mid x, A) = 0$ and thus $p(c \mid x, a) = 0$. These requirements are clearly contradictory, and so the two influences may not occur together.□

As a result there is no way that changes to 1 and 0 can conflict since the influences that cause them cannot affect the same node. This reduces the possible cases of conflict between the influences, and all the legal combinations of induced change in the probability of a node are given in Table 4. With this table we can solve the problem of conflicting influences. Consider the version of recent events according to my mother (Figure 4) who believes that as soon as I became ill, it was inevitable that I would end up in hospital. Thus, for her, knowing that I was ill immediately outweighed all the hard work I had put in taking exercise, and $\Delta p(hospital) = [\bar{\uparrow}]$.

However, landmarks do not solve every problem. It is easy to imagine real situations in which conflicting evidence will cause a problem for QP/CNs which are free of categorical influences. In qualitative reasoning circles the realised inadequacy of landmarks led to the development of 'order of magnitude' techniques, and these are what we propose to apply next.

Table 4: New sign addition

| $\oplus$ | [↑̄] | [+] | [0] | [−] | [↓̲] | [?] |
|---|---|---|---|---|---|---|
| [↑̄] | [↑̄] | [↑̄] | [↑̄] | [↑̄] |   | [?] |
| [+] | [↑̄] | [+] | [+] | [?] | [↓̲] | [?] |
| [0] | [↑̄] | [+] | [0] | [−] | [↓̲] | [?] |
| [−] | [↑̄] | [?] | [−] | [−] | [↓̲] | [?] |
| [↓̲] |   | [↓̲] | [↓̲] | [↓̲] | [↓̲] | [?] |
| [?] | [?] | [?] | [?] | [?] | [?] | [?] |

## 4   RELATIVE MAGNITUDES

The problem of conflicting evidence is precisely the sort of problem that order of magnitude systems such as ROM[K] (Dague 1993) were designed to overcome. ROM[K] is based on the idea that the order of magnitude of two quantities, $Q_1$ and $Q_2$, is usually expressed in terms of their relative sizes. Within ROM[K] there are four possible ways of expressing this relation: $Q_1$

| | | | |
|---|---|---|---|
| (A1) | $A \approx A$ | (A9) | $A \sim 1 \rightarrow [A] = [+]$ |
| (A2) | $A \approx B \rightarrow B \approx A$ | (A10) | $A \ll B \leftrightarrow B \approx (B + A)$ |
| (A3) | $A \approx B, B \approx C \rightarrow A \approx C$ | (A11) | $A \ll B, B \sim C \rightarrow A \ll C$ |
| (A4) | $A \sim B \rightarrow B \sim A$ | (A12) | $A \approx B, [C] = [A] \rightarrow (A + C) \approx (B + C)$ |
| (A5) | $A \sim B, B \sim C \rightarrow A \sim C$ | (A13) | $A \sim B, [C] = [A] \rightarrow (A + C) \sim (B + C)$ |
| (A6) | $A \approx B \rightarrow A \sim B$ | (A14) | $A \sim (A + A)$ |
| (A7) | $A \approx B \rightarrow C.A \approx C.B$ | (A15) | $A \not\approx B \leftrightarrow (A - B) \sim A$ or $(B - A) \sim B$ |
| (A8) | $A \sim B \rightarrow C.A \sim C.B$ | | |
| | | (P35) | $A \not\approx B \rightarrow C.A \not\approx C.B$ |
| (P3) | $A \ll B \rightarrow C.A \ll C.B$ | (P36) | $A \not\approx B, C \ll A \rightarrow C \ll (A - B)$ |
| (P26) | $A \sim B \rightarrow B \sim A$ | (P38) | $A \not\approx B, C \approx A, D \approx B \rightarrow C \not\approx D$ |

Figure 5: Some of the axioms and properties of ROM[K]

is *negligible wrt* $Q_2$, $Q_1 \ll Q_2$, $Q_1$ is *distant from* $Q_2$, $Q_1 \not\approx Q_2$, $Q_1$ is *comparable to* $Q_2$, $Q_1 \sim Q_2$, and $Q_1$ is *close to* $Q_2$, $Q_1 \approx Q_2$. Once the relation between pairs of quantities is specified, it is possible to deduce new relations by applying the axioms and properties of ROM[K], some of which are reproduced in Figure 5.

Together, these relations, axioms, and properties enable us to solve our on-going problems of competing influences and hypotheses by further refining the language of QP/CNs. To do this we must start dealing with the magnitude of the probabilities and influences. We denote the magnitude of the change in the probability of $A$ as $|\Delta p(A)|$, and the magnitude of the influence between $A$ and $C$ as $|S(A,C)|$, and express their relative magnitudes using the relations of ROM[K], noting that again symmetry is lost so that relations change between causal and evidential directions. Then, provided that we have a QP/CN in which the relative magnitude of the influences is known, we can apply the rules in Figure 5 to establish the relative sizes of changes at nodes of interest. Thus:

**Property 6 (relative magnitude)** Given $|S(A,C)|$ $rel_1$ $|S(B,D)|$, and $|\Delta p(A)|$ $rel_2$ $|\Delta p(B)|$, where $rel_1$, $rel_2 \in \{\approx, \sim, \not\approx, \ll\}$, then $|\Delta p(C)|$ $rel_3$ $|\Delta p(D)|$ *is given by Table 5 and the obvious symmetrical results.*

**Proof:** The change at $C$ is $|\Delta p(A)| \cdot |S(A,C)|$, and likewise for that at $D$. Thus we need the relative magnitude of the products. (i) For $|\Delta p(A)| \ll |\Delta p(B)|$ and $|S(A,C)| \ll |S(B,D)|$, the result comes from P3 and P11. (ii) For $|\Delta p(A)| \ll |\Delta p(B)|$ and $|S(A,C)| \not\approx |S(B,D)|$, we apply P3, P35 and P36 to get $|\Delta p(A)|\cdot|S(B,D)|+|\Delta p(B)|\cdot|S(A,C)| \ll |\Delta p(B)|\cdot |S(B,D)|$. Since we already know (from P3) that $|\Delta p(A)| \cdot |S(A,C)| \ll |\Delta p(A)| \cdot |S(B,D)|$, this gives us $|\Delta p(A)| \cdot |S(A,C)| + |\Delta p(B)| \cdot |S(A,C)| \ll |\Delta p(B)| \cdot |S(B,D)|$ from which the result follows. (iii) For $|\Delta p(A)| \ll |\Delta p(B)|$ and $|S(A,C)| \sim |S(B,D)|$, the result comes from A8, A11 and P3. (iv) For $|\Delta p(A)| \ll |\Delta p(B)|$ and $|S(A,C)| \approx |S(B,D)|$, the result comes from A6, A7, A11 and P3. (v) For $|\Delta p(A)| \not\approx |\Delta p(B)|$ and $|S(A,C)| \not\approx |S(B,D)|$, the result comes from the fact that $\not\approx$ is deliberately not transitive so that no relation between the products can be established. (vi) For $|\Delta p(A)| \not\approx |\Delta p(B)|$ and $|S(A,C)| \sim |S(B,D)|$, we have the same explanation. (vii) For $|\Delta p(A)| \not\approx |\Delta p(B)|$ and $|S(A,C)| \approx |S(B,D)|$, the result comes from A7, P35 and P38. (viii) For $|\Delta p(A)| \sim |\Delta p(B)|$ and $|S(A,C)| \sim |S(B,D)|$, the result comes from A8 and A5. (ix) For $|\Delta p(A)| \sim |\Delta p(B)|$ and $|S(A,C)| \approx |S(B,D)|$, the result comes from A5, A6, A7 and A8. All other results may be obtained by symmetry $\square$.

Table 5: How to establish $rel_3$ (Property 6)—$U$ indicates that the relation may not be established.

| | | $rel_2$ | | |
|---|---|---|---|---|
| | $\approx$ | $\sim$ | $\not\approx$ | $\ll$ |
| $rel_1$ $\approx$ | $\approx$ | $\sim$ | $\not\approx$ | $\ll$ |
| $\sim$ | $\sim$ | $\sim$ | $U$ | $\ll$ |
| $\not\approx$ | $\not\approx$ | $U$ | $U$ | $\ll$ |
| $\ll$ | $\ll$ | $\ll$ | $\ll$ | $\ll$ |

This result allows us to do two things. Firstly, it enables us to propagate the effect of evidence in a QP/CN so that we can distinguish which of several competing hypotheses is most strongly supported by given evidence. Consider Figure 6 which gives another version of recent events, and ponder what happens when I lose my job. The influence of losing the job on being ill is much smaller than the influence of losing the job on not being qualified, $S(lose\ job, ill) \ll S(lose\ job, not\ qualified)$, and since $|\Delta p(lose\ job)| \approx |\Delta p(lose\ job)|$ we can use Property 6 with $rel_1$ as $\ll$ and $rel_2$ as $\approx$ to find that $rel_3$ must be $\ll$. Thus $|\Delta p(ill)| \ll |\Delta p(not\ qualified)|$, and we know that the change in $p(ill)$ is much less than $p(not\ qualified)$.

Secondly Property 6 allows us to establish the effect of two competing pieces of information. If $B$ influences $C$ rather than $D$, then the relation given by Table 5 is that between the change in $p(c)$ induced by the change in $p(a)$, and that induced by the change in $p(b)$. When

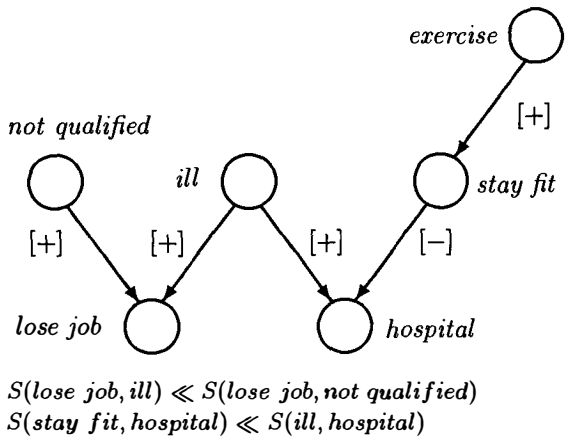

$S(lose\ job, ill) \ll S(lose\ job, not\ qualified)$
$S(stay\ fit, hospital) \ll S(ill, hospital)$

Figure 6: Some recent events (IV)

the influences compete the changes are in opposite directions, and immediately we have:

**Property 7 (comparison)** *If we have $|S(A,C)|$ rel$_1$ $|S(B,D)|$, and $|\Delta p(A)|$ rel$_2$ $|\Delta p(B)|$, where rel$_1$, rel$_2 \in \{\approx, \sim, \not\approx, \ll\}$, and $|S(A,C)|$ and $|S(B,D)|$ have opposite signs, then $[\Delta p(C)]$ is given by Table 5 and the obvious symmetrical results.*

Table 6: How to establish the sign of the change at $C$ (Property 7).

|  |  | rel$_2$ | | | |
|---|---|---|---|---|---|
|  |  | $\approx$ | $\sim$ | $\not\approx$ | $\ll$ |
| rel$_1$ | $\approx$ | [?] | [?] | $[\Delta p(B)]$ | $[\Delta p(B)]$ |
|  | $\sim$ | [?] | [?] | [?] | $[\Delta p(B)]$ |
|  | $\not\approx$ | $[\Delta p(B)]$ | [?] | [?] | $[\Delta p(B)]$ |
|  | $\ll$ | $[\Delta p(B)]$ | $[\Delta p(B)]$ | $[\Delta p(B)]$ | $[\Delta p(B)]$ |

To see how this property may be used, consider Figure 6 once again. Given that the influence of being ill on going to hospital is much greater than the influence of staying fit on not going to hospital ($|S(stay\ fit, hospital)| \ll |S(ill, hospital)|$), and that there is a roughly equal increase in the probability of my staying fit (due to knowledge of my exercising) and being ill ($|\Delta p(stay\ fit)| \approx |\Delta p(ill)|$) we can predict that there is an increase in the probability of my going to hospital when I become ill ($[\Delta p(hospital)] = [\Delta p(ill)]$).

Thus using ROM[K] allows us to solve the problem of competing influences in any situation where relative magnitude information is available—clearly many more situations than possess categorical links—and so improves on the results obtained by distinguishing truth. However, ROM[K] does not have enough numerical information to fully distinguish between competing hypotheses, only being able to predict which hypothesis undergoes the greatest change in probability. To tell which hypothesis becomes most likely when influences are not categorical we must turn to absolute order of magnitude methods.

## 5  ABSOLUTE MAGNITUDES

A suitable method for absolute order of magnitude reasoning, which revolves around the propagation of interval probability values, is discussed by Dubois et al. (1992) in the context of quantified syllogistic reasoning. In this section we adapt it to fit QP/CNs. We start by identifying suitable interval values for both influences between nodes, changes at nodes, and the values at nodes. Here we use a very basic set for reasons of brevity—more complex sets could be used if desired—and provide each with a label. The label of an interval is merely a means of referring to it, there is no claim that it is a natural linguistic interpretation of the interval. For the influences we have intervals corresponding to 'Strongly Positive' ($SP$), 'Weakly Positive' ($WP$), 'Zero' ($Z$), 'Weakly Negative' ($WN$) and 'Strongly Negative' ($SN$):

| $SP$ | $WP$ | $Z$ | $WN$ | $SN$ |
|---|---|---|---|---|
| $(1, \alpha]$ | $[\alpha, 0)$ | $0$ | $(0, -\alpha]$ | $[-\alpha, -1)$ |

where the open intervals explicitly do not allow the modelling of categorical influences. Note that, once again, these influences are not symmetrical. For both changes and values we have 'Complete Positive' ($CP$), 'Big Positive' ($BP$), 'Medium Positive' ($MP$), 'Little Positive' ($LP$) and 'Zero' ($Z$):

| $CP$ | $BP$ | $MP$ | $LP$ | $Z$ |
|---|---|---|---|---|
| $1$ | $(1, 1-\beta]$ | $[1-\beta, \beta]$ | $[\beta, 0)$ | $0$ |

The definitions of 'Little Negative' ($LN$), 'Medium Negative' ($MN$), 'Big Negative' ($BN$) and 'Complete Negative' ($CN$) are symmetrical. When propagating absolute order of magnitude quantities, we multiply change by influence to get Table 7 The ?s in this ta-ble arise because the results of these combinations depend upon the comparative values of $\alpha$ and $\beta$, and the combination of influences and changes is thus "almost-robust" (Dubois et al. 1992). If we take $\alpha \geq \beta$ and $1 - \beta \geq \alpha$, which are reasonable since typical values (to create equally wide intervals) would be $\beta \approx 0.33$ and $\alpha \approx 0.5$, then we have Table 8. Results of combining with negative influences and changes can be obtained by symmetry. where intervals such as $[MP, LP]$ in-dicate that the result falls somewhere in the interval created by the outer bounds of the named intervals.

Table 7: Combining the absolute values of influences and changes.

|  | CP | BP | MP | LP | Z |
|---|---|---|---|---|---|
| SP | ? | ? | [MP, LP] | LP | Z |
| WP | ? | ? | ? | LP | Z |
| Z | Z | Z | Z | Z | Z |

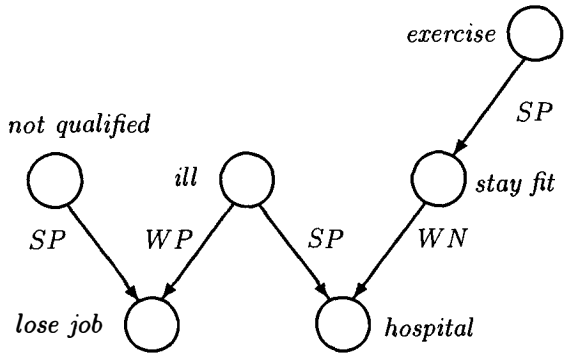

Figure 7: Some recent events (V)

Table 9: Combining the absolute values of changes and priors.

|    | CP | BP      | MP      | LP      | Z  |
|----|----|---------|---------|---------|----|
| CP |    |         |         |         | CP |
| BP |    |         |         | [CP, BP]| BP |
| MP |    |         | [CP, MP]| [CP, MP]| MP |
| LP |    | [CP, BP]| [CP, MP]| [BP, LP]| LP |
| Z  | CP | BP      | MP      | LP      | Z  |
| LN |    | [BP, MP]| [MP, Z] | [LP, Z] |    |
| MN |    | [BP, Z] | [MP, Z] |         |    |
| BN |    | [LP, Z] |         |         |    |
| CN |    |         |         |         |    |

Table 8: Combining the absolute values of influences and changes with $\alpha \geq \beta$ and $1 - \beta \geq \alpha$.

|    | CP       | BP       | MP       | LP | Z |
|----|----------|----------|----------|----|---|
| SP | [BP, MP] | [BP, MP] | [MP, LP] | LP | Z |
| WP | [MP, LP] | [MP, LP] | [MP, LP] | LP | Z |
| Z  | Z        | Z        | Z        | Z  | Z |

If we then provide for the comparison of intervals, for instance by $\geq_{int}$ (Parsons 1995b) where $[a, b] \geq_{int} [c, d]$ iff $a \geq c$ and $b \geq d$, and negation by mapping across zero into the symmetric interval, we can resolve conflicting influences. For example, consider Figure 7 in which the influences have now been given absolute orders of magnitude. Consider what happens when it is known that I both exercise and become ill. Taking into account the size of the priors, we have $\Delta p(exercise) = MP$ and $\Delta p(ill) = BP$. Using Table 8 gives $\Delta p(stay\ fit) = [MP, LP]$ so that we can calculate the effect of staying fit on going to hospital as $\Delta p(hospital)_{stay\ fit} = [[MN, LN], LN] = [MN, LN]$. The other influence on the probability of going to hospital is being ill, and clearly $\Delta p(hospital)_{ill} = [BP, MP]$. Since $|[BP, MP]| \geq_{int} |[MN, LN]|$, the biggest effect on the probability of going to hospital is being ill, and $\Delta p(hospital) = [+]$.

To resolve competing hypotheses, we need to be able to combine changes and prior values, and with our set of intervals we get Table 9. Prior values are given across the top, changes down the side. Again the results are almost robust, with those given based upon the same values of $\alpha$ and $\beta$ as before. Note that only certain combinations are possible; where they are not, the corresponding triple of prior, change, and influence cannot occur together. Now, if we are given the absolute value of the prior probabilities of the competing hypotheses, we can resolve their competition. For example, consider that the prior values of $p(lose\ job)$ and $p(hospital)$ are both $LP$, then, if $\Delta p(exercise) = MP$ and $\Delta p(not\ qualified) = BP$, then $\Delta p(hospital) = [MN, LN]$ and $\Delta p(lose\ job) = [MP, LP]$. We can use this information along with Table 9 to determine the posterior values $p^*(lose\ job) = [CP, LP]$ and $p^*(hospital) = [LP, Z]$ which by application of $\geq_{int}$ tells us that it is more likely that I will lose my job than go to hospital.

Thus the use of absolute orders of magnitude provides a solution to both the problem of competing hypotheses, and that of competing influences. Note that this method may be implemented either by the use of precompiled tables as discussed here, or more flexibly and less efficiently by the direct use of interval arithmetic.

## 6 SUMMARY

This paper has discussed various means of refining qualitative probabilistic reasoning to make it less susceptible to the problems of choosing between competing hypotheses, and of predicting the effect of conflicting influences. The first method we considered was the identification of extreme probabilities, and the categorical influences that cause such values to arise. This solved both problems, but only in the special case in which hypotheses are affected by a categorial influence. To provide more general results we used relative order of magnitude reasoning to give a good solution to the problem of conflicting influences. However, the relative method did not fully solve the problem of conflicting hypotheses, and so an absolute order of magnitude scheme was introduced. This gave a satisfactory solution to both problems.

These different schemes provide a battery of methods for extending QP/CNs which can be employed when the basic QP/CN framework is not sufficiently expressive. Clearly refining the representation will increase computational complexity, and the right degree of refinement will be determined by the particular situation to which the methods are being applied. The greater degrees of refinement are sufficient to make QP/CNs similar in scope to the $\kappa$-calculus (Darwiche & Goldszmidt 1994). While lack of space precludes a detailed comparison we can briefly point out three basic differences between the systems. Firstly, even when fully refined, QP/CNs are mainly concerned with changes in probabilities rather than probabilities themselves, unlike the $\kappa$-calculus. Secondly, QP/CNs do not require the use of infinitesimals in order to be consistent with probability theory, and so could be considered a more correct approach. Finally, the $\kappa$-calculus achieves order-of-magnitude reasoning by defining an absolute scale from which different values are taken, in contrast to the purely relative method, based on ROM[K] that is introduced here. Thus, while the $\kappa$-calculus is very similar to the absolute order of magnitude scheme introduced in Section 5, it is rather different to the other refinements discussed in this paper.

## Acknowledgements

Thanks to Ann Radda for financial support when I lost my job through illness, to John Fox for making available facilities at the Imperial Cancer Research Fund, and to Kathy Laskey, Marek Druzdzel, and the anonymous referees for helpful comments.